\documentclass{article}
\usepackage{PRIMEarxiv}
\usepackage[utf8]{inputenc}
\usepackage[T1]{fontenc}
\usepackage{hyperref}
\usepackage{url}
\usepackage{booktabs}
\usepackage{amsmath,amssymb,amsfonts}
\usepackage{nicefrac}
\usepackage{microtype}
\usepackage{fancyhdr}
\usepackage{graphicx}
\newcommand{\rev}[1]{#1}
\graphicspath{{figs/}}
\title{How memory can affect collective and cooperative behaviors in an LLM-Based Social Particle Swarm
\thanks{\rev{This is a revised version. The additional information for clarity of the model was added. The initial version also reported experiments with Gemma~3:4b but are omitted here pending further validation.}}
}
\author{
  Taisei Hishiki \\
  Graduate School of Informatics\\
  Nagoya University\\
  Nagoya, Japan\\
  \texttt{hishiki.taisei@nagoya-u.jp}
  \And
  Takaya Arita \\
  Graduate School of Informatics\\
  Nagoya University\\
  Nagoya, Japan\\
  \texttt{arita@nagoya-u.jp}
  \And
  Reiji Suzuki \\
  Graduate School of Informatics\\
  Nagoya University\\
  Nagoya, Japan\\
  \texttt{reiji@nagoya-u.jp}
}
\begin{document}
\maketitle
\begin{abstract}
This study examines how memory shapes the collective and cooperative dynamics of Large Language Model (LLM) agents in a multi-agent system.
To this end, we extend the Social Particle Swarm (SPS) model, in which agents move in a two-dimensional space and play the Prisoner's Dilemma with neighboring agents, by replacing its rule-based agents with LLM agents endowed with Big Five personality scores and varying memory lengths.
Using Gemini~2.0~Flash, we find that memory length is a critical parameter governing collective behavior: even a minimal memory drastically suppressed cooperation, transitioning the system from stable cooperative clusters through cyclical formation and collapse of clusters to a state of scattered defection as memory length increased.
Big Five personality traits correlated with agent behaviors in partial agreement with findings from experiments with human participants, supporting the validity of the model.
This effect of memory appeared whether or not personality was assigned. With heterogeneous personalities, individual behavior reflected the assigned traits and cooperation collapsed under long memory, whereas without personality Gemini's cooperative disposition dominated and cooperation was broadly maintained.
Sentiment analysis of agents' reasoning texts showed that the model interprets memory increasingly negatively as its length grows, already in the early phase, providing a micro-level account of the suppression of cooperation.
These results suggest that how an LLM interprets accumulated memory is a key driver of emergent social behavior in Generative Agent-Based Modeling.
\end{abstract}
\keywords{Large language model \and Cooperative behavior \and Social particle swarm \and Prisoner's dilemma \and Generative agent-based modeling \and Memory}
\section{Introduction}
Generative Agent-Based Modeling (GABM) using Large Language Models (LLMs) has provided novel methodologies for studying complex social phenomena by simulating agents with human-like reasoning~\cite{Park2023,Chen2023,Lu2024}.
A central question is how cooperation emerges and evolves within these AI-agent populations, connecting the rich body of evolutionary game theory~\cite{Nowak2006} to the dynamics of novel artificial agents~\cite{Sun2025}.
The ability of LLMs to exhibit strategic behavior in game-theoretical settings has been demonstrated in several studies.
Akata et al.\ found that GPT-4 behaves like a trigger strategy in repeated Prisoner's Dilemma games, always defecting after a single defection by the opponent~\cite{Akata2023}.
Fontana et al.\ reported that LLMs tend to be more cooperative than human players, attributing this to a generous behavioral heuristic instilled through alignment training~\cite{Fontana2024}.
More recently, Pal et al.\ characterized the strategies of five frontier models in iterated Prisoner's Dilemma and identified model-specific strategic profiles~\cite{Pal2026}.
These findings show that LLMs can exhibit strategy-like behavior in game-theoretic settings, motivating their use as decision-making agents.

\begin{figure}[t]
  \centering
  \includegraphics[width=0.8\columnwidth]{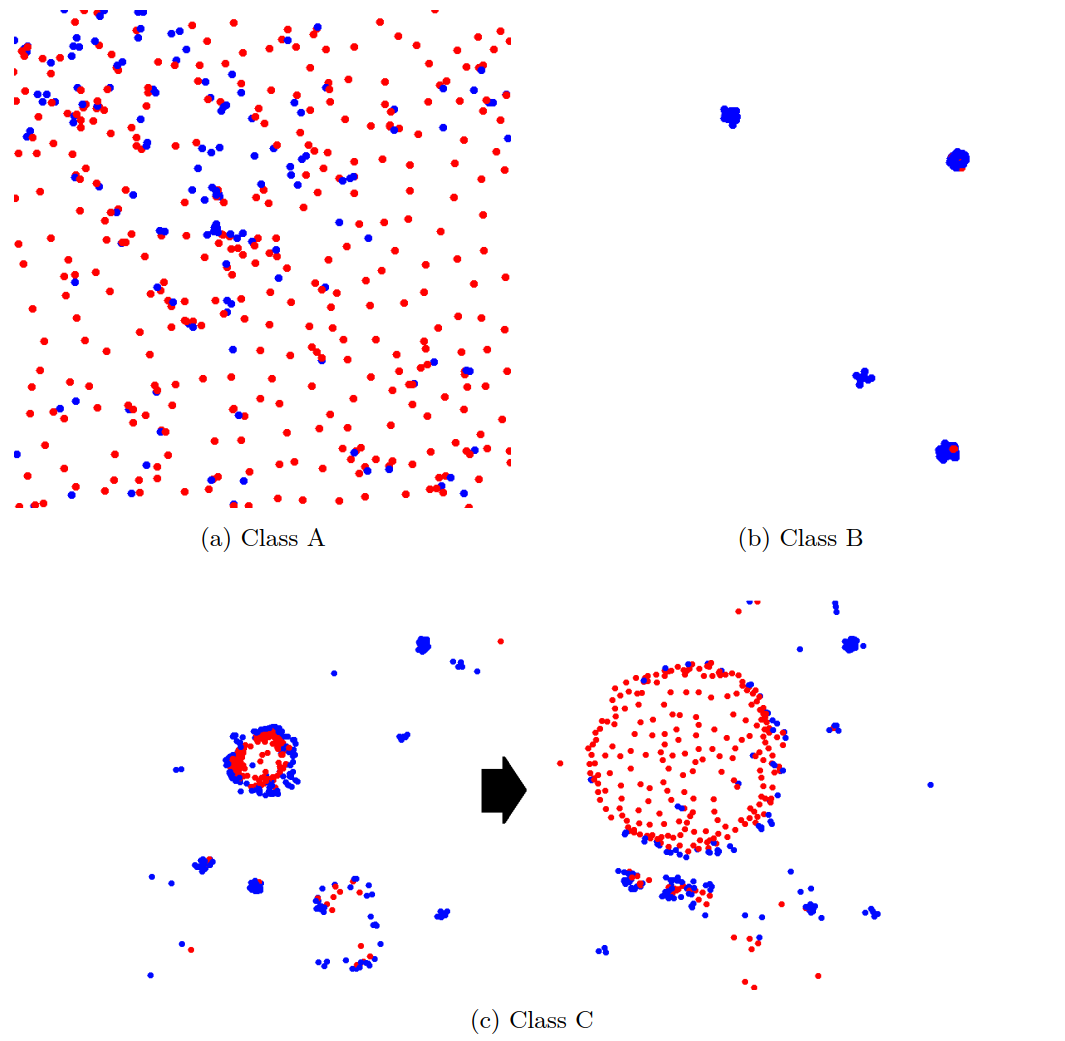}
  \caption{\textbf{Three classes of the SPS model.} (a)~Class~A: scattered and wandering defectors; (b)~Class~B: multiple and stable cooperative clusters; (c)~Class~C: repeated occurrences of emergence and collapse of cooperative clusters. Blue circles represent cooperating agents; red circles represent defecting agents.}
  \label{fig:classes}
\end{figure}

A key cognitive parameter in agent-based and mathematical models of collective behavior is memory: the ability of an agent to recall and use past interactions when making decisions.
However, the relationship between memory length and cooperative behavior presents deeply contradictory findings in the literature.
Several studies indicate that memory supports cooperation: reciprocal strategies sustain cooperation across a range of memory sizes~\cite{Hauert1997}, longer-memory strategies are more evolutionarily stable and can promote cooperation even in harsh dilemmas~\cite{Li2014,Stewart2016,Hilbe2017}, and information about the past suppresses invasions by defectors, including in spatially structured populations~\cite{Danku2019,AlonsoSanz2009}.
Others, however, report that longer memory can be detrimental, trapping agents in punishment cycles that prevent forgiveness, or that its effect is non-monotonic, peaking at an intermediate length~\cite{Horvath2012,Luo2016,Ma2021}.
Still other work finds the sign of the effect to depend on the temptation to defect, the type of dilemma, and the population structure~\cite{Qin2008,Wang2014,Murase2023}.
No unified explanation for these contradictions has emerged.

A critical limitation shared by these prior studies is that the relationship between memory and behavior is predefined: agents use memory according to fixed rules or equations, leaving no room for flexible, context-dependent interpretation.
LLM-based agents offer a qualitatively different approach.
Because LLMs process memory as natural language and reason about it within the game-theoretical context, how memory is interpreted and translated into behavior can emerge bottom-up from the agent's internal model, rather than being specified top-down by the modeler.
This raises the possibility that a history's effect on cooperation is not fixed, but depends on the internal disposition of the interpreting model.

To investigate how memory shapes cooperation in such agents, we employ the Social Particle Swarm (SPS) model~\cite{Nishimoto2023} as our experimental framework.
The SPS model integrates self-driven particle dynamics with evolutionary game theory in a continuous two-dimensional space, capturing the co-evolution of cooperative behavior and social relationships (represented as spatial distances).
It primarily exhibits three characteristic collective states: stable cooperative clusters (Class~B), cyclical formation and collapse of cooperative clusters (Class~C), and scattered defection (Class~A), as illustrated in Figure~\ref{fig:classes}.
Class~C dynamics, in particular, closely resemble the instability of cooperation observed in experiments with human participants using the SPS framework~\cite{Suzuki2018}.
Those experiments also revealed correlations between Big Five personality traits~\cite{Goldberg1981} and cooperative and spatial behavior, a finding reproduced and further developed via LLM-based personality modeling~\cite{Jiang2024,Serapio2023,Phelps2023,Suzuki2025}, which motivates the use of personality-endowed LLM agents in the present study.
Furthermore, Suzuki and Arita showed emergence and collapse of cooperative behaviors arising from evolution of personality traits of LLM agents~\cite{Suzuki2024}.

This study examines how memory length affects the cooperative dynamics of LLM-based SPS agents, showing that the LLM's interpretation of accumulated memory shapes the resulting collective behavior.
To achieve this, we extend the SPS model by replacing its rule-based agents with LLM agents whose decision-making is driven by Big Five personality scores and interaction histories of varying length.
Specifically, this paper addresses the following research questions:
(1)~How does memory length affect collective dynamics in Gemini-based SPS agents, and how do these dynamics correspond to the behavioral classes of the original model?
(2)~How do Big Five personality traits correlate with individual behavioral tendencies, and to what extent do these correlations match findings from experiments with human participants?
(3)~Does the assignment of personality traits itself, as opposed to leaving them unspecified, shape the collective dynamics?
(4)~Can the analysis of agents' natural language reasoning texts reveal the micro-level cognitive processes underlying the observed effect of memory on cooperation?
\section{Model}
The model is based on the SPS model~\cite{Nishimoto2023}, with agent decision-making replaced by LLMs.
A population of $N$ agents operates within a two-dimensional toroidal plane of size $W \times W$, interacting with others within a radius~$R$.
\subsection{Agent Behavior}
\begin{figure}[t]
  \centering
  \includegraphics[width=\columnwidth]{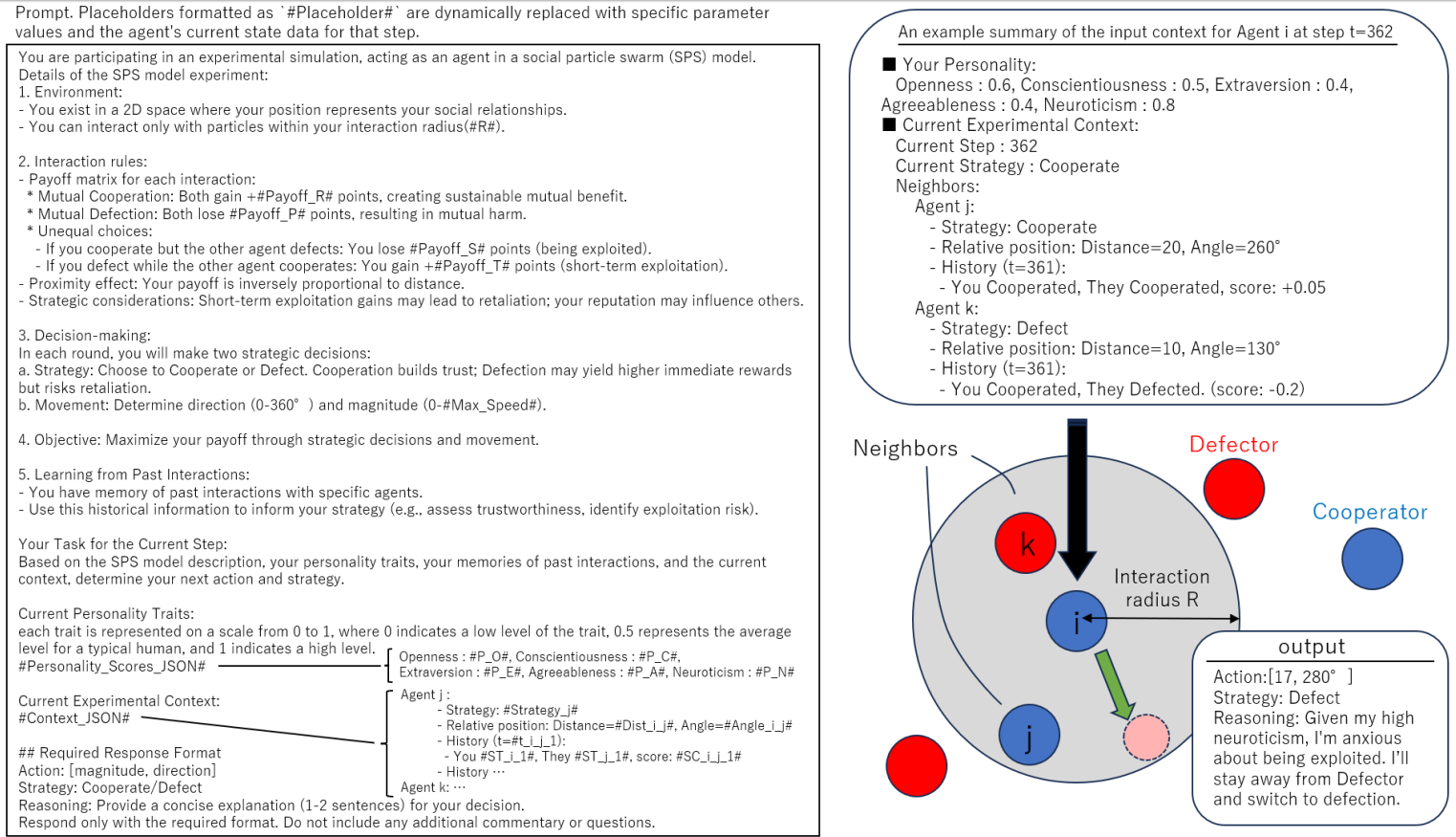}
  \caption{\textbf{Schematic overview of the LLM agent's decision-making process.} The left panel shows the overall structure of the prompt template; the individual components are described in detail in the text. The right panel shows a concrete example for agent $i$ at a specific time step: the top-right box presents the input data (current state, personality traits, interaction history, and neighborhood), the lower diagram visually represents the spatial configuration, and the output box shows the agent's final decision (strategy, movement, and reasoning).}
  \label{fig:schematic}
\end{figure}

At each time step $t$, each agent executes the following steps (see Figure~\ref{fig:schematic}):
\subsubsection{Situation recognition and decision-making.}
Each agent $i$'s LLM instance receives a prompt containing the agent's current state (position $x_i(t)$, strategy $s_i(t)$, cumulative score $\mathrm{Score}_i(t)$), its pre-assigned Big Five personality traits, its recent interaction history (based on memory length $L_m$), and the status of neighboring agents $\mathcal{N}_i(t)$ (strategy $s_j(t)$ and relative position $\mathbf{u}_{ij}(t)$).
The LLM determines the strategy $s_i(t+1)$ for the next time step and a movement action (magnitude and direction, capped at $\mathrm{MAX\_SPEED}$), outputting them together with a natural language reasoning statement.
All agents decide synchronously from the configuration at time $t$; the resulting strategies and positions are then applied together to form the configuration at $t+1$.
\subsubsection{Movement and state update.}
The position is then updated as
\begin{equation}
  x_i(t+1) = \big(x_i(t) + \Delta\mathbf{r}_i(t)\big) \bmod W,
  \label{eq:move}
\end{equation}
where $\Delta\mathbf{r}_i(t)$ is the movement vector (magnitude $\le \mathrm{MAX\_SPEED}$) and the modulo enforces the toroidal boundary. The strategy is set to $s_i(t+1)$, flipped with a small mutation probability $p_{\mathrm{mut}}$ as in the original SPS model.

\subsubsection{Payoff calculation and score update.}
Payoffs are then computed on the updated configuration, with $g_{\mathrm{base}}$ following the standard ordering $T > R > P > S$ (values given below).
The instantaneous total score $G_i(t)$ is given by
\begin{equation}
  G_i(t) = \sum_{j \in \mathcal{N}_i(t)}
            \frac{g_{\mathrm{base}}(s_i(t),\, s_j(t))}
                 {1 + |\mathbf{u}_{ij}(t)|},
  \label{eq:payoff}
\end{equation}
where $g_{\mathrm{base}}(s_i, s_j)$ is the basic payoff of the Prisoner's Dilemma game between agents $i$ and $j$, and the denominator penalizes payoffs for spatially distant neighbors.
The score is then updated as
\begin{equation}
  \mathrm{Score}_i(t+1) = \mathrm{Score}_i(t) + G_i(t).
  \label{eq:score}
\end{equation}
\subsection{Prompt Design}
As schematically illustrated in Figure~\ref{fig:schematic} (left panel), the prompt provided to each agent at every time step is structured into five components.
Placeholders of the form \texttt{\#Variable\#} are dynamically replaced with the agent's actual values at each step.
\subsubsection{Game settings and objective}
The prompt opens by defining the experimental environment: agents exist in a two-dimensional space where position represents social relationships, and interaction is limited to neighbors within radius $R$.
It then specifies the full Prisoner's Dilemma payoff matrix, including the proximity effect by which payoffs are inversely weighted by distance between agents.
The agent's objective is stated as ``Maximize your cumulative payoff,'' without prescribing any specific tactic.
This design is intentional: the trade-off between short-term exploitation and long-term trust-building is left entirely to the LLM's own reasoning, allowing strategic behavior to emerge bottom-up from the agent's internal model.
\subsubsection{Personality traits}
Each agent's Big Five personality scores~\cite{Goldberg1981} (Openness, Conscientiousness, Extraversion, Agreeableness, and Neuroticism) are provided in JSON format, as shown in Figure~\ref{fig:schematic} (right panel, ``Current Personality Traits'').
Each trait is expressed as a continuous value in $[0, 1]$, accompanied by a note explaining that 0 indicates a low level of the trait, 0.5 represents the average level for a typical human, and 1 indicates a high level.
This quantitative representation provides a replicable means of assigning individuality, in contrast to purely textual persona descriptions.
For example, an agent with a high Agreeableness score is expected to adopt a cooperative bias, while an agent with a high Neuroticism score is expected to be sensitive to the risk of exploitation and to respond defensively to adverse experiences.
\subsubsection{Current experimental context}
The agent's own current state (position, strategy, and cumulative score) and the status of all neighbors within radius $R$ are provided in JSON format, as shown in Figure~\ref{fig:schematic} (right panel, ``Current Experimental Context'').
The relative position of each neighbor is expressed in polar coordinates (distance and angle centered on the focal agent) rather than Cartesian coordinates, so that the LLM can intuitively interpret spatial proximity and direction.
Each neighbor's current strategy (Cooperate or Defect) is also included, allowing the agent to assess the current distribution of strategies in its neighborhood.
\subsubsection{Interaction history (memory)}
When $L_m > 0$, the interaction history with each neighbor is appended in JSON format to that neighbor's entry in the current experimental context, as shown in Figure~\ref{fig:schematic} (right panel, ``History'').
The history records the $L_m$ most recent interactions with that opponent, each entry containing the time step at which the interaction occurred, the strategies adopted by both agents, and the payoff received by the focal agent.
The records are sorted in reverse chronological order so that the most recent interaction appears first.
This opponent-specific history enables context-dependent decisions such as retaliating against a past defector or maintaining cooperation with a trusted partner.
The history is kept per opponent and persists even when that opponent leaves the interaction radius.
For $L_m = 0$, no history is included and the agent decides solely on the basis of the current state.
Because the LLM is stateless and retains no internal memory across API calls, the entire relevant history is injected as text into the prompt at every time step.
\subsubsection{Output format}
The agent is instructed to respond in a structured format specifying three fields: \textit{Action}, \textit{Strategy}, and \textit{Reasoning}.\footnote{If a response could not be parsed into this format, the agent remained in place and its strategy defaulted to cooperation. Such parse failures were rare with Gemini (about $0.006\%$ of decisions in the personality-endowed experiments and under $0.7\%$ in the unspecified-personality condition) and did not affect the results.}
The \textit{Action} field specifies the movement vector as a magnitude $\in [0, \mathrm{MAX\_SPEED}]$ and a direction $\in [0^\circ, 360^\circ]$, as illustrated in Figure~\ref{fig:schematic} (right panel, ``output'').
The \textit{Strategy} field specifies the next cooperative or defective action.
The \textit{Reasoning} field elicits a concise natural language rationale (one to two sentences) explaining the decision in terms of the agent's personality and past interactions.
This reasoning output is not merely supplementary: it enables direct analysis of the micro-level cognitive processes underlying macro-level collective behavior, as exploited in the sentiment analysis described later.
\subsubsection{Illustrative example}
To make the above concrete, consider the example shown in Figure~\ref{fig:schematic} (right panel).
Agent $i$ is at time step $t = 362$ and is currently cooperating.
Its personality profile includes a high Neuroticism score ($= 0.8$), indicating a strong sensitivity to the risk of exploitation.
In the neighborhood, there are two agents: Agent $j$ (cooperating, distance~20, angle~260\textdegree) and Agent $k$ (defecting, distance~10, angle~130\textdegree).
The interaction history shows that in the previous step ($t = 361$), Agent $j$ cooperated with Agent $i$ (payoff $+0.05$), while Agent $k$ defected against Agent $i$ (payoff $-0.2$).
Based on this situation, the LLM outputs Action $[17, 280\textdegree]$ and Strategy: Defect, with the reasoning: ``Given my high neuroticism, I am anxious about being exploited. I will stay away from the defector and switch to defection.''
This example illustrates how the agent integrates personality traits, spatial context, and interaction history to produce context-dependent behavior: the high Neuroticism score amplifies the negative experience with Agent $k$, leading to a defensive switch to defection and a movement away from the defector.
\subsection{Language Model Used}
Gemini~2.0~Flash (Google) is a commercial model optimized for interactive use and safety-aligned for general users via extensive reinforcement learning from human feedback.
In this study, Gemini~2.0~Flash serves as the decision-making engine for all agents.
\section{Experiments and Results}
\subsection{Experimental Settings}
The experimental parameters were set as follows: $N = 100$, $W = 500$, $R = 50$, $\mathrm{MAX\_SPEED} = 20$, $T = 500$ steps per trial.
The Prisoner's Dilemma payoff matrix used the following values: $Payoff_T$ (temptation) $= 2.0$, $Payoff_R$ (reward) $= 1.0$, $Payoff_P$ (punishment) $= -1.0$, $Payoff_S$ (sucker's payoff) $= -2.0$.
At the start of each trial, each agent's Big Five trait scores were drawn independently from a truncated normal distribution with mean~$0.5$, standard deviation~$0.16$, clipped to $[0,1]$.
Positions were initialized uniformly at random and initial strategies set to cooperate or defect with equal probability; the decoding temperature was $0.7$ and the mutation probability $p_{\mathrm{mut}} = 0.03$.
We used Gemini~2.0~Flash as the decision-making language model.
Memory length $L_m \in \{0, 1, 2, 3\}$ was varied across conditions, with 10 independent trials per condition.
Experimental codes, data, and videos of typical dynamics are publicly available online~\cite{data2025}.
\subsection{Effect of Memory Length on Collective Dynamics}
\label{subsec:gemini}
Table~\ref{tab:gemini} summarizes the mean and volatility (average of within-trial standard deviations) of the cooperation rate and average neighbor count under each $L_m$ condition.
As $L_m$ increased, the mean cooperation rate decreased monotonically from 0.899 ($L_m = 0$) to 0.0776 ($L_m = 3$), while the average number of neighbors decreased accordingly.
The highest volatility in cooperation rate occurred at $L_m = 1$, indicating significant temporal variability and dynamic social relationships at this intermediate memory length.
\begin{table}[t]
  \centering
  \caption{\textbf{Mean and volatility of the number of neighbors and cooperation rate for each $L_m$ condition (Gemini~2.0~Flash, with personality).} Volatility is the average of standard deviations calculated from the time series of each metric within a single trial, averaged over 10 trials.}
  \label{tab:gemini}
  \begin{tabular}{ccccc}
    \toprule
    & \multicolumn{2}{c}{Number of Neighbors} & \multicolumn{2}{c}{Cooperation Rate} \\
    \cmidrule(lr){2-3}\cmidrule(lr){4-5}
    $L_m$ & Mean & Volatility & Mean & Volatility \\
    \midrule
    0 & 17.6 & 6.41  & 0.899  & 0.0454 \\
    1 & 3.75 & 1.80  & 0.260  & 0.108  \\
    2 & 2.65 & 1.38  & 0.139  & 0.102  \\
    3 & 2.48 & 0.387 & 0.0776 & 0.0462 \\
    \bottomrule
  \end{tabular}
\end{table}

\begin{figure}[t]
  \centering
  \includegraphics[width=0.9\columnwidth]{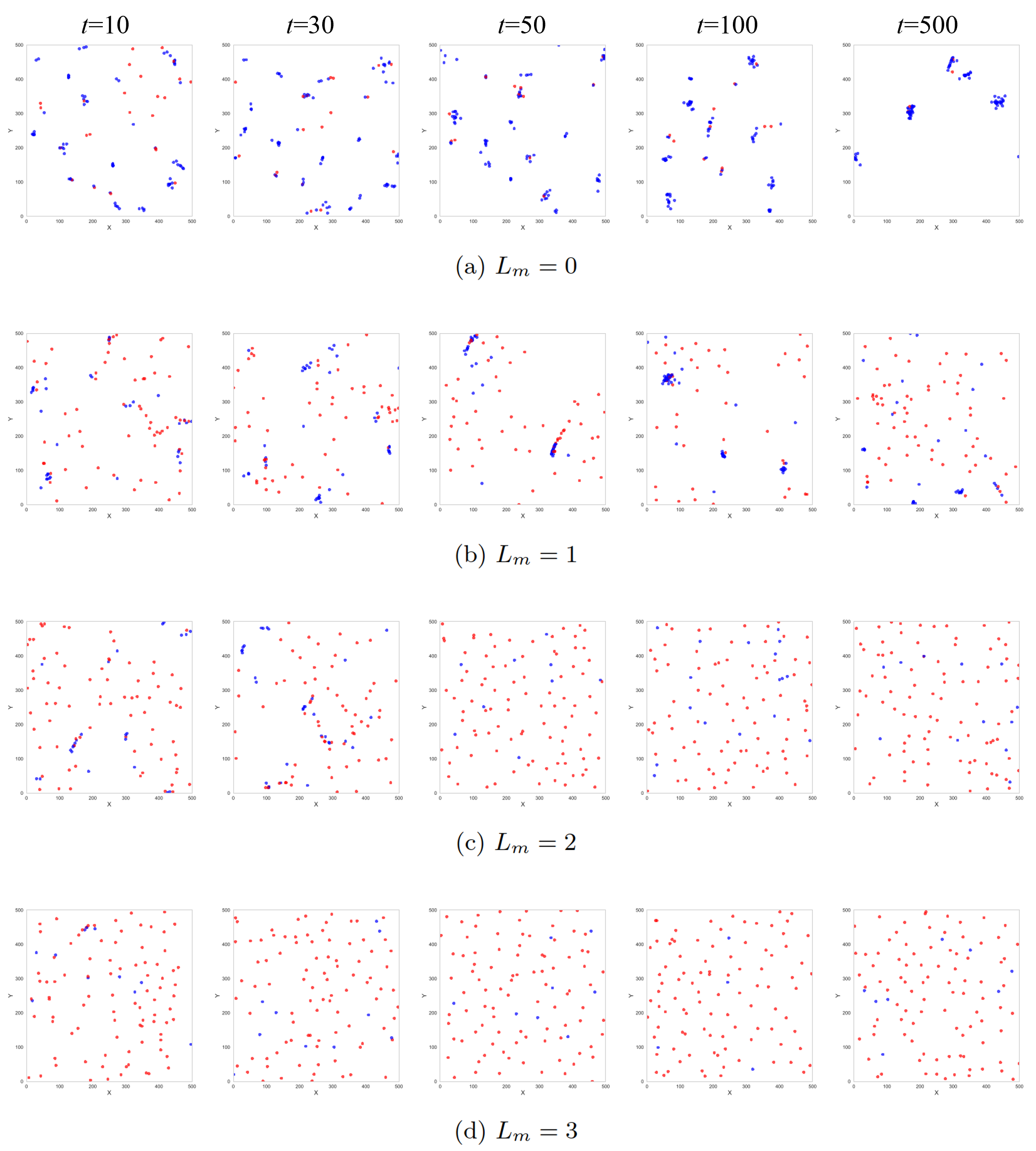}
  \caption{\textbf{Agent spatial configurations for $L_m = 0, 1, 2$, and $3$ (Gemini~2.0~Flash).} Each row shows snapshots at $t = 10, 30, 50, 100$, and $500$ for the corresponding $L_m$ condition. Blue circles represent cooperating agents; red circles represent defecting agents. (a)~$L_m = 0$: cooperative clusters form and grow (Class~B-like); (b)~$L_m = 1$: cyclical formation and collapse of clusters (Class~C-like); (c)~$L_m = 2$: gradual transition toward defection; (d)~$L_m = 3$: rapid convergence to scattered defection (Class~A-like).}
  \label{fig:snaps}
\end{figure}

As shown in Figure~\ref{fig:snaps}(a), for $L_m = 0$ the population began as a dispersed mixture of cooperators and defectors at early time steps, but cooperative clusters rapidly formed and merged over time, resulting in a large stable cluster by $t = 500$.
This is consistent with Class~B dynamics in the original SPS model.

For $L_m = 1$, Figure~\ref{fig:snaps}(b) illustrates the cyclical nature of the dynamics: cooperative clusters emerged in the early to middle phase, but were subsequently invaded and collapsed, with this cycle repeating throughout the experiment.
The wide fluctuations in both cooperation rate and neighbor count are characteristic of Class~C dynamics.

For $L_m = 2$, Figure~\ref{fig:snaps}(c) shows that while some cooperative grouping was visible in the early phase, agents became progressively more isolated as the experiment proceeded.
At $L_m = 3$, Figure~\ref{fig:snaps}(d) shows that the population rapidly dispersed into isolated defecting agents with almost no cooperative interaction remaining by $t = 500$, corresponding to Class~A dynamics.
These results demonstrate that a single cognitive parameter, memory length, can qualitatively shift the collective dynamics of the system, replicating the class transitions observed in the original SPS model.

A notable spatial pattern was also observed: defecting agents tended to drift leftward (approximately 180\textdegree), while cooperating agents tended to move toward the upper right (approximately 60\textdegree).
This directional bias might reflects an inherent tendency of the LLM to choose up-right direction (60\textdegree) when positive situations and to select a midpoint direction (180\textdegree) when no clear motivational signal is present in the neighborhood.
\subsection{Effect of Personality Traits on Individual Behavior}
\label{subsec:personality}
\begin{figure}[t]
  \centering
  \includegraphics[width=0.7\columnwidth]{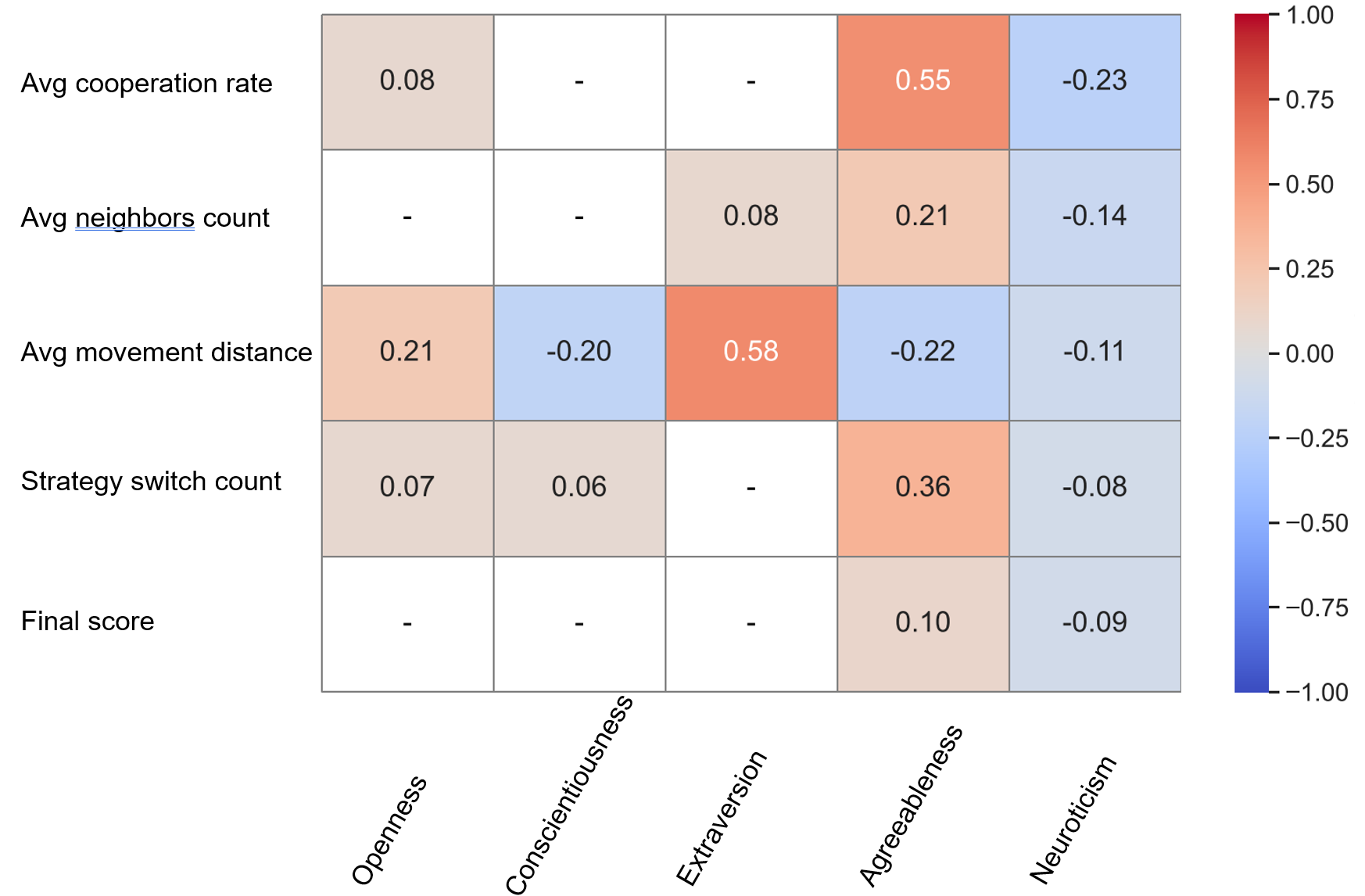}
  \caption{\textbf{Pearson correlation coefficients between Big Five personality traits and agent behavioral characteristics ($L_m = 1$ condition, Gemini~2.0~Flash).} Each cell shows the mean Pearson $r$ value averaged over 10 trials; cells marked ``-'' indicate non-significant correlations ($p \geq 0.05$). Warmer colors indicate positive correlations; cooler colors indicate negative correlations.}
  \label{fig:personality}
\end{figure}

To assess the validity of the model and examine how pre-assigned Big Five personality traits influence agent behavior, we calculated the Pearson correlation coefficient between each trait score and five behavioral metrics (cooperation rate, average neighbor count, movement distance, strategy switch count, and final score) across all agents in each trial, at $L_m = 1$.
The mean Pearson $r$ values, averaged over the 10 trials, are shown in the heatmap in Figure~\ref{fig:personality}; cells where the correlation was not significant ($p \geq 0.05$) in any trial are marked with ``-''.

Agreeableness showed the strongest and most consistent positive correlation with cooperation rate ($r = 0.55$) and with neighbor count ($r = 0.21$), and a negative correlation with movement distance ($r = -0.22$), indicating that agreeable agents cooperated more and formed stable clusters with less spatial exploration.
Extraversion correlated positively with movement distance ($r = 0.58$), consistent with explorative behavior.
Neuroticism showed a negative correlation with cooperation rate ($r = -0.23$), suggesting a risk-averse tendency leading to withdrawal rather than engagement.

These patterns are partially consistent with findings from experiments with human participants in the SPS framework reported by Suzuki et al.~\cite{Suzuki2018}, where agreeable participants formed clusters and moved less, and extraverted participants moved more.
A difference was observed for Neuroticism: whereas neurotic human participants tended to switch strategies frequently (trial and error behavior), neurotic LLM agents showed a weak negative correlation with strategy switching and withdrew spatially rather than varying their strategy.
This may reflect the LLM's tendency to minimize loss under uncertainty, rather than exploring through behavioral variation.
These results support the validity of our model, demonstrating that LLM agents express personality-consistent behavioral tendencies that partially mirror those observed in human participants.
\subsection{Effect of Removing Personality Assignment}
\label{subsec:nopersonality}
To isolate the effect of personality assignment, we ran an additional condition in which the personality section was omitted from the prompt (the ``Unspecified Personality'' condition), with all else unchanged (Table~\ref{tab:nopersonality}).
Cooperation still declined with memory length, but far less steeply: at $L_m = 1$ it remained 0.834, versus 0.260 with personality.
Even at $L_m = 3$ the population did not collapse to scattered defection (Class~A) but retained Class~C-like cyclical dynamics.
\begin{table}[t]
  \centering
  \caption{\textbf{Mean and volatility of the number of neighbors and cooperation rate for each $L_m$ condition without personality assignment (Gemini~2.0~Flash, ``Unspecified Personality'' condition).} Volatility is defined as in Table~\ref{tab:gemini}, averaged over 3 independent trials.}
  \label{tab:nopersonality}
  \begin{tabular}{ccccc}
    \toprule
    & \multicolumn{2}{c}{Number of Neighbors} & \multicolumn{2}{c}{Cooperation Rate} \\
    \cmidrule(lr){2-3}\cmidrule(lr){4-5}
    $L_m$ & Mean & Volatility & Mean & Volatility \\
    \midrule
    0 & 15.8 & 8.27  & 0.962 & 0.0292 \\
    1 & 7.12 & 1.24  & 0.834 & 0.0717 \\
    2 & 7.58 & 1.46  & 0.750 & 0.0821 \\
    3 & 7.12 & 0.996 & 0.509 & 0.0963 \\
    \bottomrule
  \end{tabular}
\end{table}

Memory thus strongly shapes the collective dynamics in both cases, but the outcome depends on individuality.
With assigned personalities, behavior reflects the individual traits (Section~\ref{subsec:personality}), and defection-prone agents (e.g., low Agreeableness or high Neuroticism) seed the distrust that lets long memory drive the population to full defection.
Without personalities, Gemini's cooperative disposition dominates and the population stays broadly cooperative, consistent with the original rule-based SPS model, in which a homogeneous population converges on a cooperative cluster~\cite{Nishimoto2023}.
\subsection{Analysis of Memory Interpretation via Reasoning Texts}
\label{subsec:sentiment}
To investigate the micro-level cognitive processes underlying the memory-driven suppression of cooperation, we analyzed the natural language reasoning statements produced by each agent.
From each agent's reasoning text, we extracted sentences containing memory-related keywords (``memory,'' ``remember,'' ``past,'' ``history,'' ``previous,'' ``last,'' ``before,'' ``ago,'' ``earlier,'' ``former'').
Sentiment analysis was then applied to the extracted sentences using a pre-trained DistilBERT model fine-tuned on the SST-2 dataset~\cite{Sanh2020} (\texttt{distilbert-base-uncased-finetuned-sst-2-english} from Hugging Face).
The model outputs a sentiment score in $[-1, +1]$, where positive values indicate a positive emotional tone and negative values indicate a negative tone.
We treat this score as a proxy for whether the agent interprets its memory positively (as a basis for cooperation) or negatively (as a basis for caution and defection).

Table~\ref{tab:sentiment_full} (``Full'' column) shows the average sentiment scores per $L_m$ condition for Gemini~2.0~Flash over the full experiment ($t = 0$--$500$). At $L_m = 0$, no interaction history is provided to the agent; nevertheless, memory-related keywords still appear in the reasoning texts, presumably because agents refer to the current neighborhood state as a basis for anticipating future interactions. We therefore treat the $L_m = 0$ scores as a baseline that reflects the model's intrinsic sentiment tendency in the absence of explicit interaction history.
The score was positive at $L_m = 0$ (approximately $+0.60$) but declined sharply with increasing memory length, falling into negative territory at $L_m = 3$ (approximately $-0.24$).

\begin{table}[t]
  \centering
  \caption{\textbf{Average sentiment scores of memory-related sentences in Gemini~2.0~Flash agents' reasoning texts.} Mean sentiment score ($-1$: negative, $+1$: positive) of memory-related reasoning sentences, per memory length $L_m$, for the full experiment ($t = 0$--$500$) and the early phase ($t \leq 30$). In both cases the score is positive at $L_m = 0$ and declines into negative territory at $L_m = 3$. The consistency of this pattern in the early phase indicates that the increasingly negative memory interpretation is an intrinsic property of the model and not an artifact of the converged social state.}
  \label{tab:sentiment_full}
  \begin{tabular}{ccc}
    \toprule
    $L_m$ & Full ($t = 0$--$500$) & Early ($t \leq 30$) \\
    \midrule
    0 & $+0.60$ & $+0.49$ \\
    1 & $+0.06$ & $+0.24$ \\
    2 & $-0.02$ & $+0.21$ \\
    3 & $-0.24$ & $-0.09$ \\
    \bottomrule
  \end{tabular}
\end{table}

Because the sentiment scores could reflect the macro-level state of the population rather than an intrinsic memory-interpretation tendency of the model, we repeated the analysis restricting to the early phase of the experiment ($t \leq 30$), before the population-level cooperation dynamics had converged.
The ``Early'' column of Table~\ref{tab:sentiment_full} shows the results for this early phase.
The score remained positive at $L_m = 0$ (approximately $+0.49$) but again declined with increasing memory length, falling into negative territory at $L_m = 3$ (approximately $-0.09$).
This negative shift is already present in the early phase, indicating it is a genuine property of the model rather than an effect of the converged state.

Qualitatively, the reasoning texts mirror this shift: at $L_m = 0$ the agents' rationales are dominated by prospective, trust- and relationship-building language, whereas with longer memory they increasingly cite past defections and adopt a retrospective, distrust- and exploitation-focused framing.

These results provide a micro-level explanation for the macro-level suppression of cooperation.
The model interprets accumulated memory increasingly negatively: past defections trigger a defensive, risk-averse response that suppresses cooperation.
This corresponds to the ``punishment trap'' of Horvath et al.~\cite{Horvath2012}, in which accumulated records of past defections lock agents into mutual retaliation.
More generally, how an LLM interprets accumulated memory can determine whether memory promotes or suppresses cooperation, offering a lens on the mixed findings of prior work.
\section{Conclusion}
We investigated how interaction histories (memory) shape cooperative behavior and collective dynamics in an LLM-based Social Particle Swarm model.
Using Gemini~2.0~Flash, we found that memory length is a critical parameter: increasing it drove the system from stable cooperative clusters, through cyclical formation and collapse, to scattered defection, reproducing the behavioral classes of the rule-based SPS model.
Big Five traits correlated with individual behavior in partial agreement with human-participant experiments, supporting the model's validity.
Memory strongly affected the dynamics whether or not personality was assigned; the difference lay in how.
With personalities, behavior reflected the individual traits and cooperation collapsed under long memory, whereas without them Gemini's cooperative disposition dominated and the population stayed broadly cooperative.

Sentiment analysis traced this suppression to an increasingly negative interpretation of accumulated memory. This collapse corresponds to the punishment trap of classical models~\cite{Horvath2012}, suggesting that LLM agents can recapitulate mechanisms long studied in formal models of cooperation.

These findings suggest that how an LLM interprets accumulated memory is a key determinant of emergent social behavior in GABM, understandable in terms of the agent's internal model rather than the game rules alone.

More broadly, as LLM agents are increasingly deployed in multi-agent systems and automated workflows, the social dynamics that emerge from such systems may be governed less by the explicit rules of interaction and more by how the underlying model interprets its inputs.
Understanding and characterizing these behavioral tendencies is therefore a prerequisite for designing reliable and predictable LLM-based societies.

Future work will include introducing explicit reasoning phases prior to action selection, extending memory representation to qualitative natural-language impressions, and conducting a broad comparison across LLMs, including larger open-weight models, to characterize how model-specific properties shape memory interpretation.
\section{Data Availability}
The codes and data that support the findings of this study, including videos of typical dynamics under different experimental conditions, are publicly available via the figshare repository at \url{https://figshare.com/s/5fc0ff99b469e0d21256}.
\section{Acknowledgements}
This study was supported in part by JSPS Topic-Setting Program to Advance Cutting-Edge Humanities and Social Sciences Research Grant Number JPJS00122674991, JSPS KAKENHI 24K15103, and the Google Gemma~2 Academic Program.
\bibliographystyle{unsrt}
\bibliography{references}
\end{document}